\documentclass[letterpaper, 10 pt, conference]{ieeeconf}  % Comment this line out if you need a4paper

\IEEEoverridecommandlockouts                              % This command is only needed if 
\usepackage{cite}
\usepackage{amsmath,amssymb,amsfonts}
\usepackage{algorithmic}
\usepackage{graphicx}
\usepackage{textcomp}
\usepackage{xcolor}
\usepackage{orcidlink}
\usepackage{times}
\usepackage{epsfig}
\usepackage{amssymb}
\usepackage{flexisym}
\usepackage{booktabs}
\usepackage{multirow}
\usepackage{array}
\usepackage{tabularx}
\usepackage{ragged2e}
\usepackage{multicol}
\usepackage{subfigure}
\usepackage{gensymb}
\usepackage{tabstackengine}
\usepackage{adjustbox}
\usepackage{etoolbox}
\usepackage{bm}
\usepackage{xurl}
\usepackage[all]{nowidow}
\usepackage[ruled,vlined]{algorithm2e}
\usepackage{booktabs}
\usepackage{amsmath}
\usepackage{pifont}
\usepackage{siunitx}

\title{\LARGE \bf
CollisionSplatting: Collision-Aware Motion Planning in 3DGS Scenes with Image-Conditioned Objectives and Adjustable Conservatism
}
\author{R. Khorrambakht$^{1}$, Joaquim Ortiz-Haro$^{1}$, Stephan Weiss$^{3}$, and Ludovic Righetti$^{1,2}$
\thanks{$^{1}$ Machines in Motion Lab, Tandon School of Engineering, New York University}
\thanks{$^{2}$ Artificial and Natural Intelligence Toulouse Institute (ANITI)}
\thanks{$^{3}$ Department of Smart Systems Technologies, University of Klagenfurt, Austria.}
\thanks{Code is available at \url{https://github.com/machines-in-motion/CollisionSplatting}.}
}

\DeclareMathAlphabet\mathbfcal{OMS}{cmsy}{b}{n}
\mathcode`*=\string"8000
\begingroup
\catcode`*=\active
\xdef*{\noexpand\textup{\string*}}
\endgroup

\begin{document}

\maketitle

\begin{abstract}
Incorporating dense visual information into motion planning remains challenging, as geometric planners rely on abstracted scene representations that discard visual richness, while learned visual models often lack geometric interpretability and computational efficiency. This paper introduces CollisionSplatting, a simple, modular, GPU-accelerated, probability-inspired distance metric with tunable conservatism that operates directly on standard 3D Gaussian Splatting (3DGS) scenes. When combined with learned image-conditioned reward functions, this metric enables joint geometric and visual planning by unifying collision-aware costs with image-space objectives. We integrate the metric into GPU-accelerated Model Predictive Path Integral (MPPI) and Rapidly-Exploring Random Tree (RRT) planners, and show on-par or better collision-classification performance compared to representative baselines while achieving substantially higher collision-checking throughput and significantly lower VRAM usage. Finally, we demonstrate the effectiveness of our metric in real-world vision-guided navigation and manipulation tasks, highlighting 3DGS as a practical bridge between rich perception and real-time motion planning.
\end{abstract}

\IEEEpeerreviewmaketitle
\section{Introduction}
\label{sec:intro}
Reducing visual information into primitive geometrical abstractions has been an effective approach to systematically isolating perception and planning subsystems in robotics. This separation, however, comes at the expense of information bottlenecks and reduced robustness. 3D Gaussian Splatting (3DGS) has recently emerged as an intermediate representation that preserves rich visual information while explicitly capturing the 3D geometry of the scene. In fact, this representation and its rasterizer can be regarded as an action-conditioned world model: given a candidate future robot pose, the rasterizer predicts the corresponding visual observation at a fraction of the cost of large learned world models~\cite{bar2025navigation, lu2025gwm}.

Motivated by this potential, we introduce \emph{CollisionSplatting}, a highly efficient and modular distance/collision metric that operates directly on the standard 3DGS representation and can be seamlessly incorporated into standard motion planning and control algorithms as a single-kernel GPU module. We show that, when combined with visual reward models, our metric enables real-time collision-aware planning while respecting semantic objectives imposed by the reward model.

Most existing work proposes distance metrics that are tightly coupled to a particular planning algorithm or robot model. Splat-Nav~\cite{DBLP:journals/corr/abs-2403-02751} was among the earliest and introduced a robot–Gaussian binary collision classifier as the core of a safe-flight-corridor-based planning system. Splat-Nav replaces each Gaussian in 3DGS with an equivalent probability-level-set ellipsoid and computes its collision state with the robot, which is likewise modeled as an ellipsoid.  SAFER-SPLAT~\cite{chen2024safer} also considers each Gaussian as an ellipsoid, but instead of a binary check, quantifies the distance between the robot (spherical) and each Gaussian via a constrained numerical optimization problem on top of which a Control Barrier Function (CBF) safety filter is formulated.

While Splat-Nav and SAFER-SPLAT replace Gaussians with deterministic collision primitives, ATLASNav~\cite{atlas_navigator} adopts a probabilistic view, modeling both the robot and Gaussians as spatial probability distributions. To achieve analytical tractability, ATLASNav assumes isotropic 3DGS Gaussians, spherical robot geometry, and approximately homogeneous scene density. Finally, SPLANNING~\cite{michaux2025let} introduces a statistically rigorous, risk-aware collision-avoidance framework for 3DGS-based manipulation. However, to guarantee probabilistic consistency, SPLANNING relies on a custom \emph{Normalized 3DGS} reconstruction pipeline, meaning its theoretical guarantees apply to that modified representation rather than to standard unnormalized 3DGS datasets~\cite{InteriorGS} or existing generative models~\cite{worldlabs-website}.

In contrast, our work proposes a simple distance/collision metric inspired by a probability-propagation scheme. It operates directly on standard 3DGS reconstructions and accommodates both anisotropic and isotropic Gaussians, disk-like 2DGS Gaussians~\cite{Huang2DGS2024}, and ellipsoidal or spherical multi-link robot primitives. Its simplicity and streaming feed-forward computation graph make our metric highly parallelizable, with per-splat-localized memory requirements that avoid large intermediate allocations or multi-stage GPU/CPU data transfers. Consequently, our method achieves significant memory efficiency and high collision-checking throughput while maintaining on-par or better performance compared to baselines.

Furthermore, we show how the efficiency of our metric, combined with the efficient 3DGS rasterizer, enables real-time joint visual–geometric planning under a Model Predictive Path Integral (MPPI) formulation for both a 7-DoF manipulator and a quadruped mobile robot. This MPPI integration example is closely related to BEINGS~\cite{BEINGS}, which formulates image-goal navigation in image space, but our approach explicitly penalizes collisions in real time using a single-kernel GPU-accelerated collision cost.
% based on our metric.

% The rest of the paper is organized as follows. Sec.~\ref{sec:method} briefly reviews the background of the 3DGS representation and presents the proposed distance metric. Then baseline comparison and ablation studies are presented in Sec.~\ref{sec:evaluation} followed by real-world deployment examples in Sec.~\ref{sec:demonstrations}. Finally, the paper is concluded in \ref{sec:conclusion}.

The paper is organized as follows. Sec.~\ref{sec:method} reviews the 3DGS background and presents our distance metric, Sec.~\ref{sec:evaluation} covers baseline comparisons and ablations, and Sec.~\ref{sec:demonstrations} presents real-world deployments before concluding in Sec.~\ref{sec:conclusion}.

\section{Method}
\label{sec:method}
\subsection{Background}
\label{sec:background}
Gaussian splatting may be thought of as an extension of point-clouds. It represents the 3D scene as a set of Gaussian primitives (referred to as splats in this paper) $\mathcal{S}=\{\mathcal{G}_i\}$ with locations and spans parameterized by translation vectors $\boldsymbol{\mu}^{g}_i \in \mathbb{R}^3$ and covariance matrices $\boldsymbol{\Sigma}^g_i \in \mathbb{R}^{3\times 3}$:
\begin{equation}
    \label{eq:gaussian}
    \mathcal{G}_i(\bm{p}) = \exp\!\Big(-\tfrac{1}{2}(\bm{p}-\boldsymbol{\mu}^{g}_i)^\top{\boldsymbol{\Sigma}^g_i}^{-1}(\bm{p}-\boldsymbol{\mu}^{g}_i)\Big),
\end{equation}
where $\bm{p} \in \mathbb{R}^3$ is the splat's position in the world frame $\mathcal{W}$. Each covariance $\boldsymbol{\Sigma}^g_i$ is parametrized by a diagonal matrix $\bm{S}_i$ describing the scale of each principal axis in the Gaussian's local frame $\mathcal{G}_i$ and a rotation matrix ${^{\mathcal{W}}\bm{R}_{\mathcal{G}}}_i$ that represents the orientation of this frame with respect to the global world frame $\mathcal{W}$:
\begin{equation}
    \label{eq:cov_decomposition}
    \boldsymbol{\Sigma}^g_i = {^{\mathcal{W}}\bm{R}_{\mathcal{G}}}_i\;\bm{S}_i\bm{S}_i^\top\;{^{\mathcal{W}}\bm{R}_{\mathcal{G}}}_i^\top.
\end{equation}

The differentiable rendering process includes transforming each 3D Gaussian into the camera frame $\mathcal{C}$ and then projecting it onto the image plane via a local affine map $\bm{J}$~\cite{kerbl20233d}:
\begin{equation}
    \label{eq:covariance_projection}
    \boldsymbol{\Sigma}'_{i} = \bm{J}\;{^{\mathcal{C}}\bm{R}_{\mathcal{G}}}_i\,\boldsymbol{\Sigma}^g_i\,{^{\mathcal{C}}\bm{R}_{\mathcal{G}}}_i^\top\,\bm{J}^\top.
\end{equation}
The first two rows and columns of $\boldsymbol{\Sigma}'_{i}$ yield the 2D covariance of the projected 3D Gaussian $\mathcal{G}_i$ onto the image plane. The projection of the 3D means $\boldsymbol{\mu}^{g}_i$ follows the conventional point-based projection mapping $\boldsymbol{\mu}' = \pi(\boldsymbol{\mu}^{g}_i, {^{\mathcal{C}}\bm{T}_{\mathcal{G}_i}}, \bm{K}_{\text{cam}})$, where ${^{\mathcal{C}}\bm{T}_{\mathcal{G}_i}} \in \mathbb{R}^{4\times4}$ is the homogeneous transformation from Gaussian frame $\mathcal{G}_i$ to camera frame $\mathcal{C}$ and $\bm{K}_{\text{cam}}\in\mathbb{R}^{3\times 3}$ is the camera intrinsic matrix.

Given these 2D projections of the 3D Gaussians on the image plane, the color at point $\bm{x}$ in image space is computed through volumetric alpha blending of the depth-wise sorted Gaussian primitives along the ray passing through $\bm{x}$:
\begin{equation}
\label{eq:alpha_blending}
\bm{c}(\bm{x})=\sum_{k=1}^K \bm{c}_k \alpha_k \mathcal{G}_k^{2D}(\bm{x}) \prod_{j=1}^{k-1}\left(1-\alpha_j \mathcal{G}_j^{2D}(\bm{x})\right),
\end{equation}
where $\bm{c}_k$ and $\alpha_k$ are learnable parameters representing the color and opacity of Gaussian $\mathcal{G}_k$, and $\mathcal{G}_k^{2D}(\bm{x})$ is the 2D counterpart of \eqref{eq:gaussian} parametrized by $\boldsymbol{\Sigma}'_i, \boldsymbol{\mu}'_i$ in image space. The term $\prod_{j=1}^{k-1}\left(1-\alpha_j \mathcal{G}_j^{2D}(\bm{x})\right)$ indicates how much light is occluded by other Gaussian primitives between the camera and Gaussian $\mathcal{G}_k$.

% During reconstruction, the scene is initialized randomly or using priors from SfM or RGB-D cameras. Then, an alternating densification and parameter optimization is employed to minimize the $\mathcal{L}_1$ loss between the rendered and ground-truth images. 
During reconstruction, the scene is initialized randomly or from SfM/RGB-D priors, then optimized by alternating densification and parameter updates to minimize the L1 rendering loss.
Each Gaussian primitive may be augmented with any $d$-dimensional task-dependent feature vector (similar to color). During rendering, these features are projected into the image space, creating novel-view feature maps. Depth maps can be similarly rendered by replacing the color in \eqref{eq:alpha_blending} with the splat's distance from the camera.

\subsection{CollisionSplatting Distance Metric}
\label{sec:distance_formulation}
We assume that the robot may be decomposed into a set of collision ellipsoids. Compared to previous works that assume spherical representations~\cite{jin2024gs, chen2024safer, atlas_navigator}, our choice allows for a greater level of expressivity without trading off computational efficiency.
\subsubsection{Distance Metric Between a Point and an Ellipsoid}
Consider the robot as a collection of ellipsoidal bodies $\mathcal{E}_j$, each expressed as the $s_{ij}^2=1$ level set:
\begin{equation}
\label{eq:ellipsoid}
   (\bm{X}-\boldsymbol{\mu}^{r}_j)^\top \bm{A}_j(\bm{X}-\boldsymbol{\mu}^{r}_j)= s_{ij}^2,
\end{equation}
where $\boldsymbol{\mu}^r_j \in \mathbb{R}^3$ denotes the center of the ellipsoidal link in the world frame and $\bm{A}_j$ encodes its span and orientation. The matrix $\bm{A}_j$ may be decomposed as
$\bm{A}_j = {^{\mathcal{W}}\bm{R}_{\mathcal{E}_j}} \,\bm{P}_j\, {^{\mathcal{W}}\bm{R}_{\mathcal{E}_j}}^\top$, where ${^{\mathcal{W}}\bm{R}_{\mathcal{E}_j}}$ is the orientation of the ellipsoid body frame $\mathcal{E}_j$ with respect to the world frame, and
\begin{equation}
    \bm{P}_j = \begin{bmatrix}
        1/a_j^2 & 0 & 0\\
        0 & 1/b_j^2 & 0\\
        0 & 0 & 1/c_j^2
    \end{bmatrix},
\end{equation}
with $a_j, b_j, c_j \in \mathbb{R}_+$ denoting the lengths of each semi-axis of $\mathcal{E}_j$ in its body frame. Now consider $\boldsymbol{\mu}^g_i \in \mathbb{R}^3$ as the position of a point obstacle in the world frame. Substituting $\bm{X}=\boldsymbol{\mu}^g_i$ in \eqref{eq:ellipsoid} yields $s_{ij}^2$ with three possible cases: if the point is on the surface of the body, $s_{ij}^2=1$; if it is inside, $s_{ij}^2<1$; and if the point is outside (not colliding with the ellipsoid), $s_{ij}^2>1$. This measure, often referred to as a scaling-function distance, is employed in~\cite{sailingPC} to formulate a GPU-accelerated CBF-based framework using LiDAR point clouds.

\subsubsection{Distance Metric Between a Gaussian and an Ellipsoid}
We now extend the point-based metric in \eqref{eq:ellipsoid} to work with Gaussian obstacles. We consider each Gaussian $\mathcal{G}_i$ as a probability density over the possible positions of a point obstacle $\bm{p}_i$, with $\bm{p}_i \sim \mathcal{N}(\boldsymbol{\mu}^{g}_i, \boldsymbol{\Sigma}^g_i)$.

Drawing inspiration from the uncertainty-propagation view of the 3DGS projection in Eq.~\ref{eq:covariance_projection}, we treat the point-based distance function $s_{ij}^2(\bm{X})$ as a nonlinear transformation and adopt a linearization-based covariance propagation scheme to map input uncertainty (3DGS scale) into uncertainty over the distance. Specifically, we linearize $s_{ij}^2(\bm{X})$ around $\bm{X}=\boldsymbol{\mu}^g_i$ and propagate the input uncertainty (point obstacle location) to yield a 1D Gaussian distribution $s_{ij}^2 \sim \mathcal{N}(\mu^{s}_{ij}, \sigma_{ij}^2)$, where $\mu^{s}_{ij}= s_{ij}^2(\boldsymbol{\mu}^g_i)$ and $\sigma_{ij}^2$ is computed as:
\begin{equation}
    \label{eq:linearization}
    \sigma^2_{ij} = \left ( \left. \frac{\partial s_{ij}^2}{\partial \bm{X}} \right|_{\bm{X}=\boldsymbol{\mu}^{g}_i} \right ) \boldsymbol{\Sigma}^g_i \left (\left. \frac{\partial s_{ij}^2}{\partial \bm{X}} \right|_{\bm{X}=\boldsymbol{\mu}^{g}_i} \right )^\top.
\end{equation}
This propagation is accurate when the nonlinearity of $s_{ij}^2(\cdot)$ is moderate in a neighborhood of $\boldsymbol{\mu}^g_i$ and the Gaussian scale is small relative to the robot–obstacle distance. While this assumption empirically holds in well-reconstructed 3DGS scenes, in Sec.~\ref{sec:experiments-linearize} we show that even when it is violated, our metric becomes more conservative, which is a desirable failure mode for safety-critical planning.

We then define our conservative distance metric as
\begin{equation}
    \label{eq:prob_distance_single_splat}
    \tilde{s}_{ij}^2 =  \max\big(0,\; s_{ij}^2 - n_\sigma\sqrt{\sigma^2_{ij}}\big),
\end{equation}
where $n_\sigma \geq 0$ controls the level of conservatism and can be set based on a desired confidence level. Setting $n_\sigma=0$ recovers the original point-based distance, implying that the same planning pipeline can be seamlessly used with point clouds. Finally, to determine the collision state of a robot link, we adopt a numerically stable recursive softmin operation over the splats. Note that the complexity of evaluating the collision state between a $K$-link robot and a Gaussian scene with $N$ primitives is $O(KN)$. In practice the robot has a limited reach at each instant, so we adopt a GPU-accelerated spatial hash grid to filter out unreachable parts of the map.

\subsection{Assumptions and Design Choices}
Our formulation is probability-inspired rather than statistically calibrated: the propagated variance in \eqref{eq:linearization} is treated as an uncalibrated uncertainty measure. Furthermore, we aggregate per-splat distances using local operations (recursive softmin over neighbors), which keeps computations localized and avoids global normalization or coupling between Gaussians, thereby reducing memory footprint and runtime. Proper uncertainty calibration of the aggregated distance metric remains an interesting direction for future work. In addition, 3DGS scenes may represent transparent or reflective objects using mixtures of low-opacity splats away from the true surface. For collision checking, we assume obstacles of interest are not highly transparent and discard splats that are both very low opacity and large in spatial extent to filter floaters. Modern 3DGS variants such as 2DGS~\cite{Huang2DGS2024} tend to reduce such artifacts, further mitigating the need for this heuristic.

\subsection{MPPI Controller With Image-Based Cost}\label{sec:mppi}
We adopt a standard MPPI formulation for receding-horizon control. At each iteration, we sample $K$ control sequences around a nominal sequence $\bar{U}$, roll them out in parallel under the system dynamics, and evaluate the resulting trajectories with a cost
\begin{equation}
    C^k = C_{\text{collision}}^k + C_{\text{task}}^k + C_{\text{image}}^k + C_{\text{control}}^k,
\end{equation}
where $C^k_{\text{task}}$ encodes any geometric tracking objectives, $C^k_{\text{control}}$ penalizes control effort, $C^k_{\text{collision}}$ is the collision avoidance cost, and $C^k_{\text{image}}$ represents any visual cost function. The trajectory costs are converted into weights via a softmin with temperature $\lambda$:
\begin{equation}
    \label{eq:mppi_softmin}
    w^k = \frac{\exp(-C^k / \lambda)}{\sum_{l=1}^{K}\exp(-C^l / \lambda)},
\end{equation}
and these weights are used to update the nominal sequence $\bar{U}$ in the direction of lower-cost perturbations. After a small, fixed number of iterations, the first control in the updated sequence is applied to the robot, and $\bar{U}$ is warm-started at the next MPC step by temporally shifting the previous solution. In practice, we parameterize the control sequence by B-spline knots to ensure control signal smoothness~\cite{howell2022}.

\subsubsection{Collision Cost}
We use our collision metric in \eqref{eq:prob_distance_single_splat} to define a collision cost along each trajectory. For each robot link $j$ and Gaussian $i$, we define
\begin{equation}
\label{eq:mppi_collision_cost}
d_{ij} = d_{\max}\exp\big(-\gamma \max(\tilde{s}_{ij}^2 - s_{\min}, 0)\big),
\end{equation}
where $\gamma \in \mathbb{R}_+$ is a user-defined decay factor that reduces the influence of distant splats, $d_{\max}$ is the maximum cost per splat, and $s_{\min}$ is the minimum allowable scaling factor (set to one in our case). At each time step, these per-splat, per-link values are averaged over neighboring splats (via spatial hashing) and summed over links to yield a running collision cost. The trajectory collision cost $C^k_{\text{collision}}$ is then computed as the sum of this running cost over the horizon. Our GPU implementation evaluates this term on batches of shape $(K, T, N_{\text{link}})$, where $K$ is the number of rollouts, $T$ the prediction horizon, and $N_{\text{link}}$ the number of ellipsoidal primitives describing the robot.

\subsubsection{Semantic Image-Based Cost}
\label{semantic-cost}
Gaussian splatting can efficiently render images from arbitrary viewpoints in parallel~\cite{ye2024gsplatopensourcelibrarygaussian}. We leverage this capability to add a terminal visual cost. Specifically, the terminal state of each trajectory is first fed to the 3DGS rasterizer to synthesize an image. These images are then passed through a visual encoder to evaluate a learned reward. In this paper, we formulate this reward as the squared Euclidean distance in the embedding space of a foundation visual encoder (DINOv2 in our experiments):
\begin{equation}
\left\lVert \mathrm{enc}(\bm{I}_{\text{target}}) - \mathrm{enc}(\bm{I}_G) \right\rVert_2^2,
\end{equation}
where $\mathrm{enc}(\cdot)$ denotes the image encoder, $\bm{I}_{\text{target}}$ the goal image, and $\bm{I}_G$ the image generated from the terminal state of the trajectory.

\subsection{RRT Global Planner}
We further demonstrate the applicability of our distance metric in global motion planning by implementing a batched GPU-accelerated Rapidly-Exploring Random Tree (RRT) algorithm. The collision metric in Eq.~\ref{eq:prob_distance_single_splat} is used as a parallelized collision-checking engine for all candidate edges in each batch. Specifically, we compute the distance metric $s_{ij}^2$ between each robot link and all splats, marking a state as in collision if any link has more than a small user-defined number of Gaussians with $s_{ij}^2 < 1$.

The RRT grows in parallel on the GPU. At each iteration, $N_{\text{batch}}$ random samples are drawn from the configuration space, and for each, the closest tree node is expanded toward the sample up to a maximum stride length. Each edge is checked for collisions at discrete points defined by a small user-specified resolution, and collision-free edges are appended as new nodes.

All tree data structures and buffers reside in GPU memory to minimize data transfer. Collision checking is implemented as a kernel operating on batches of shape $N_{\text{batch}} \times N_{\text{strides}} \times N_{\text{links}}$, where $N_{\text{strides}}$ denotes the number of checks per edge based on the chosen resolution.
\section{CollisionSplatting Metric Evaluation}
\label{sec:evaluation}
\subsection{Setup}
We conduct our study here using the Stonehenge scene adopted by Splat-Nav~\cite{DBLP:journals/corr/abs-2403-02751}. It features various gate-like structures and columns with varying opening widths (Fig.~\ref{fig:pf_steps}), thereby representing different degrees of navigation difficulty. The scene scale is normalized to $1$, and the robot is modeled as a single sphere of radius $0.015$ (i.e., $1.5\%$ of the scene scale) for the baseline metric evaluations. We use NVIDIA Warp\cite{warp2022} to implement our custom kernels and execute all simulated studies on a workstation equipped with an NVIDIA RTX 6000 Pro GPU and an AMD Ryzen 9 9950X CPU.

\subsection{Baselines Comparison}
In this section, we compare our distance metric against representative baselines in terms of collision classification performance and computational throughput. As representative of methods modeling Gaussians as elliptical collision primitives, we choose Splat-Nav~\cite{DBLP:journals/corr/abs-2403-02751} and SAFER-SPLAT~\cite{chen2024safer}. Additionally, we compare our metric against ATLASNav~\cite{atlas_navigator}, a closely related work that derives a probabilistic distance metric for isotropic Gaussian scenes and spherical robots (ours operates on the more general case of anisotropic 3DGS scenes and ellipsoidal robots). We also include SPLANNING~\cite{michaux2025let}, which rigorously upper-bounds the probability of collision between a robot (represented as a collection of spheres) and a radiance field by computing the volume integral of a normalized 3DGS density. 
% We note that the custom normalized 3DGS reconstruction pipeline is not open-sourced at the time of writing. We therefore evaluate the SPLANNING metric on our standard 3DGS reconstruction. However, because the precision-recall curve is generated by sweeping the probability threshold from 0 to 1 rather than relying on absolute probability values, we expect this limitation to have minimal effect on the study's fairness.
Since SPLANNING's normalized reconstruction pipeline was not open-sourced at the time of writing, we evaluate its metric on our standard 3DGS reconstruction; because the precision-recall curve sweeps the threshold from 0 to 1 rather than using absolute probabilities, this has minimal effect on fairness.
% \footnote{Conditioned on the release of the reconstruction pipeline and if required, we will update our study before camera-ready submission.}.
Furthermore, note that we focus on the distance/collision metric proposed for each baseline in this section and do not consider the planner choices to maintain comparability.

\begin{figure}[t]
    \begin{center}
        \includegraphics[width=0.9\linewidth]{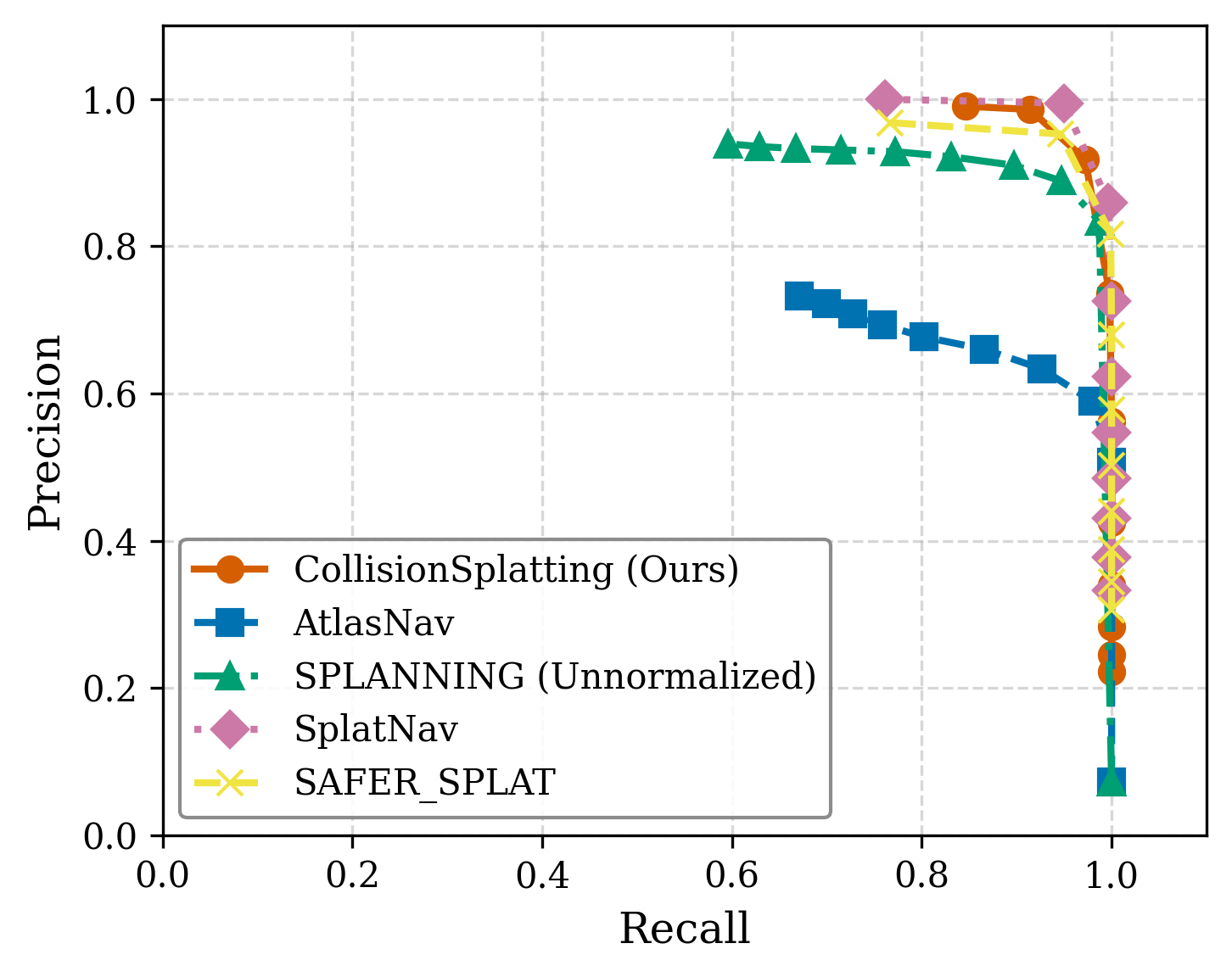}
        \caption{Precision-recall curves for CollisionSplatting distance metric and representative baselines.}
        \label{fig:pr-curve}
    \end{center}
\end{figure}

We evaluate each distance metric on a grid of points in the Stonehenge environment ($\approx \num{200000}$ Gaussians splats) and show the precision-recall curves for each method, along with the computational throughput and GPU RAM utilization in Fig.~\ref{fig:pr-curve} and Table~\ref{tab:runtime_comparison}, respectively. We see that our metric performs on par with Splat-Nav, SAFER-SPLAT, and SPLANNING, while outperforming ATLASNav.
% \footnote{ATLASNav was not publicly available at the time of writing, thus results are based on our own implementation of their metric.}. 
We attribute ATLASNav's lower performance to its isotropic Gaussian assumption. In our study, we accommodate this need by replacing the anisotropic Gaussians with spherical counterparts through scale averaging. 

In terms of computational efficiency, Table~\ref{tab:runtime_comparison} reveals that while ATLASNav and SPLANNING implementations support batched GPU evaluation, their reliance on large intermediate tensor allocations results in significant memory consumption for applications where batched collision checking is essential (e.g., zeroth-order optimization and MPPI). In our experiment, these methods respectively required 55.0~GB and 14.5~GB of GPU memory for a batch size of 4,096. In contrast, our CollisionSplatting metric evaluates collision states using a direct, single-pass, per-splat propagation technique that entirely avoids intermediate allocations. In our experiment, this streaming formulation only required 0.16~GB of VRAM for the same batch size and scales to larger batch sizes without a significant increase in memory requirements (we report batch size of $32,768$ with a negligible VRAM increase to 0.166~GB). This memory efficiency and localized in-kernel computation model enable significantly higher, more scalable throughput for our method, which directly enables our real-time experiments.

% Remember to include \usepackage{graphicx} in your preamble for \resizebox

% \begin{table}[t]
% \centering
% \caption{Runtime Performance Comparison of Collision Detection Baselines}
% \label{tab:runtime_comparison}
% \resizebox{\linewidth}{!}{% <-- The % prevents an extra space from causing overfull hboxes
% \begin{tabular}{l c c c c}
% \toprule
% \multirow{2}{*}{\textbf{Method}} & \textbf{Runtime (GPU)} & \textbf{Runtime (CPU)} & \textbf{Acceleration} & \multirow{2}{*}{\textbf{Batched}} \\
%  & \textbf{mean $\pm$ STD (s)} & \textbf{mean $\pm$ STD (s)} & $\mathbf{(t_{CPU}/t_{GPU})}$ & \\
% \midrule
% SplatNav & $17.9 \pm 0.11$ & $91.4 \pm 0.74$ & $5.1\times$ & \xmark \\
% SAFER-SPLAT & $17.4 \pm 0.17$ & $82.7 \pm 0.39$ & $4.75\times$ & \xmark \\
% Atlas-Nav & $0.251 \pm 0.006$ & $7.5 \pm 0.03$ & $29.8\times$ & \cmark \\
% \textbf{CollisionSplatting (Ours)} & $\mathbf{0.011 \pm 0.002}$ & $\mathbf{0.18 \pm 0.003}$ & $16.4\times$ & \cmark \\
%  & & $(0.71 \pm 0.004)^{\dagger}$ & $(\mathbf{64.5\times})^{\dagger}$ & \\
% \bottomrule
% \multicolumn{5}{l}{\footnotesize $^{\dagger}$ Collision check on batches of 16,384 robot poses instead of 4096.}
% \end{tabular}% <-- The % here is also helpful
% }
% \end{table}

\begin{table*}[t]
\centering
\caption{Collision Detection Throughput and VRAM Utilization Comparison}
\label{tab:runtime_comparison}
\resizebox{\textwidth}{!}{
\begin{tabular}{l c c c c c}
\toprule
\multirow{2}{*}{\textbf{Method}} & \textbf{Throughput (GPU) ($\uparrow$)} & \textbf{Throughput (CPU) ($\uparrow$)} & \textbf{Acceleration ($\uparrow$)} & \textbf{VRAM Utilization ($\downarrow$)} & \multirow{2}{*}{\textbf{Batch Size}} \\
 & \textbf{Checks/s (mean $\pm$ STD)} & \textbf{Checks/s (mean $\pm$ STD)} & $\mathbf{(n_{\text{GPU}}/n_{\text{CPU}})}$ & \textbf{(GB)} & \\
\midrule
SplatNav~\cite{DBLP:journals/corr/abs-2403-02751} & $162.1 \pm 1.6$ & $23.7 \pm 0.5$ & $6.84\times$ & $0.22$ & 1 \\
SAFER-SPLAT~\cite{chen2024safer} & $217.5 \pm 3.4$ & $34.4 \pm 0.5$ & $6.33\times$ & $0.38$ & 1 \\
ATLASNav$^{\dagger}$~\cite{atlas_navigator} & $16,318.7 \pm 381.0$ & $262.2 \pm 17.3$ & $62.23\times$ & $55.0$ & $4,096$ \\
SPLANNING~\cite{michaux2025let} & $157,538.5 \pm 603.6$ & $2,409.4 \pm 18.3$ & $65.38\times$ & $14.50$ & $4,096$ \\
\midrule
\multirow{2}{*}{\textbf{CollisionSplatting (Ours)}} & $215,578.9 \pm 7,660.2$ & $13,653.3 \pm 440.4$ & $15.7\times$ & \textbf{0.16} & $4,096$ \\
 & $\mathbf{1,638,400.0 \pm 32,125.5}$ & $\mathbf{12,750.2 \pm 88.7}$ & $\mathbf{128.5\times}$ & \textbf{0.166} & $32,768$ $(4,096\times8)$ \\
\bottomrule
\multicolumn{6}{l}{\footnotesize $^{\dagger}$ Evaluated using our own implementation, as the official source code for ATLASNav was not publicly available at the time of writing.}
\end{tabular}
}
\end{table*}

\subsection{Ablation Study}
We implement a GPU-accelerated RRT planner that leverages the efficiency and speed of our collision metric to grow the tree 32 nodes at a time. We further optimize the global path from RRT by implementing a zeroth-order optimizer that minimizes path length while penalizing collisions (a single MPPI step, as described in Sec.\ref{sec:mppi}, with a horizon set to the full plan length). We evaluate this planner in the Stonehenge scene and on a set of randomized trials, in which the robot starts at the center and navigates to a randomly selected goal along the scene's border. 

\subsubsection{Conservatism Factor} 
\begin{figure*}[!t]
    \centering
    % ---- Row 1 ----
    \subfigure[$n_\sigma=0$]{\includegraphics[width=0.17\textwidth]{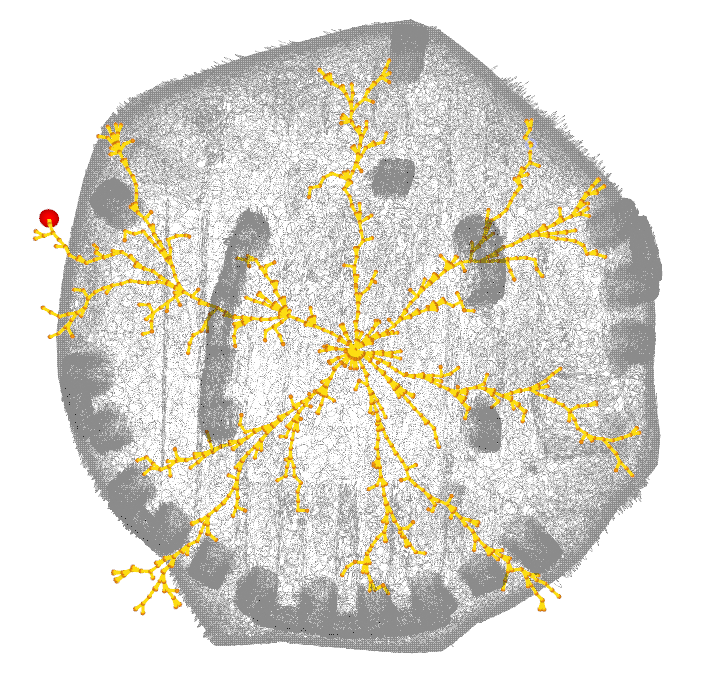}}
    \subfigure[$n_\sigma=0.5$]{\includegraphics[width=0.17\textwidth]{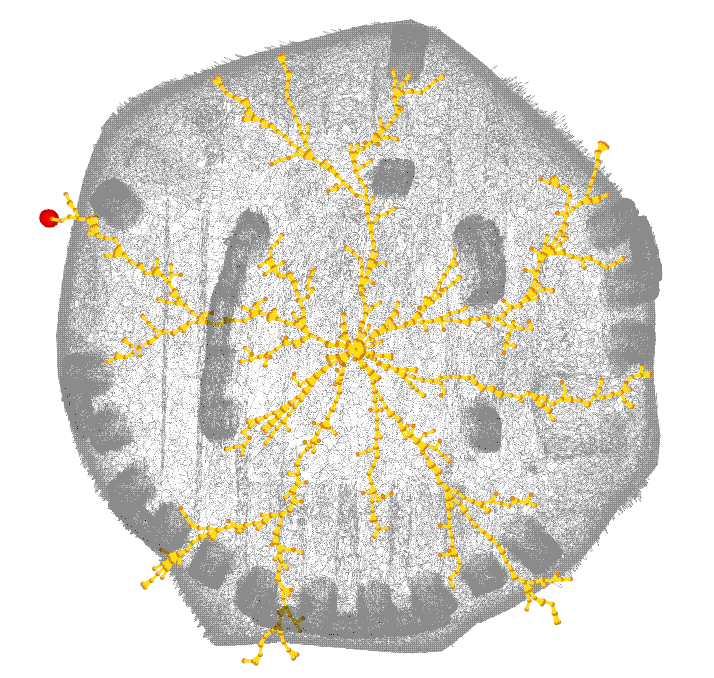}}
    \subfigure[$n_\sigma=1$]{\includegraphics[width=0.17\textwidth]{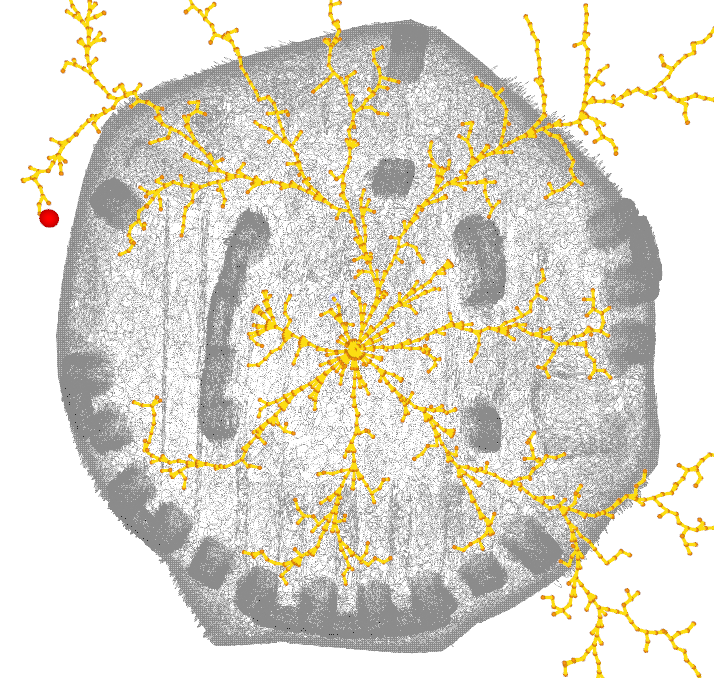}}
    \subfigure[$n_\sigma=1.5$]{\includegraphics[width=0.17\textwidth]{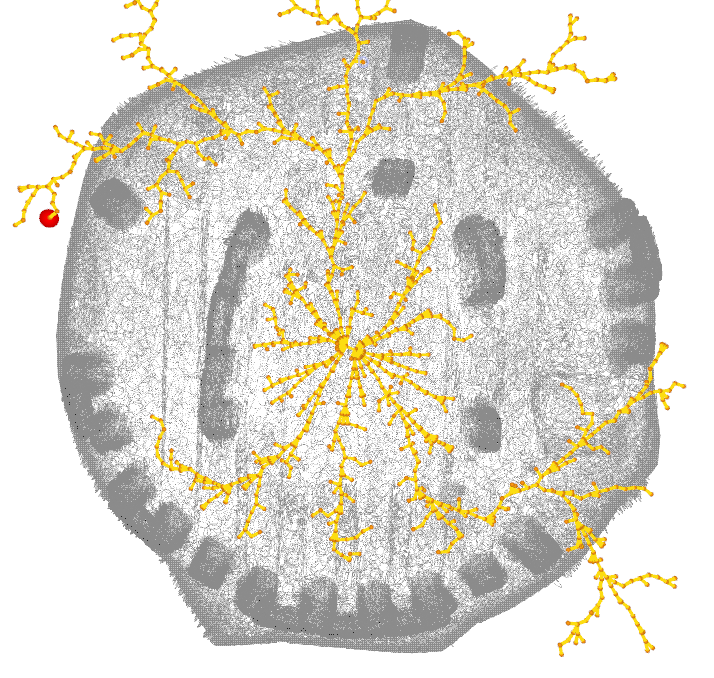}}

    \caption{Effect of the conservatism factor $n_\sigma$ on the paths discovered by the RRT planner. Increasing $n_\sigma$ yields more conservative trajectories that avoid tight passages.}
    \label{fig:pf_steps}
\end{figure*}

Table~\ref{tab:exp1-tab1} reports the dimensionless path length (normalized scene), the minimum distance to the closest Gaussian mean $d_{\min}$ (a proxy for safety), and normalized path smoothness for different conservatism factor $n_\sigma$ values before and after refinement. Note that when $n_\sigma=0$, the uncertainty propagation is nullified, and the scene effectively reduces to a dense point-cloud of Gaussian means, implying that our metric may also be seamlessly applied to point-cloud maps. We can see that as $n_\sigma$ increases, both the path length and the minimum obstacle distance grow, indicating more conservative plans. This effect is also shown qualitatively in Fig.~\ref{fig:pf_steps}, where larger $n_\sigma$ values cause the planner to avoid narrow gates and navigate around obstacles instead. Notably, after refinement, obstacle-avoiding clearances increase while the path length decreases, and since optimization occurs directly over B-spline knots, the resulting paths are also smooth by design.

\begin{table*}
\caption{Global RRT planning and local MPPI refinement vs various conservatism factors.}
\label{tab:exp1-tab1}
\resizebox{\textwidth}{!}{
\begin{tabular}{ccccccc}
\hline
\begin{tabular}[c]{@{}c@{}}$n_{\sigma}$\\ (conservatism)\end{tabular} & \begin{tabular}[c]{@{}c@{}}Path Length  ($\downarrow$)\\ (RRT)\end{tabular} & \begin{tabular}[c]{@{}c@{}}Path Length ($\downarrow$)\\ (RRT+Refinement)\end{tabular} & \begin{tabular}[c]{@{}c@{}}$d_{min}$ ($\uparrow$)\\ (RRT)\end{tabular} & \begin{tabular}[c]{@{}c@{}}$d_{min}$ ($\uparrow$)\\ (RRT+Refinement)\end{tabular} & \begin{tabular}[c]{@{}c@{}}Smoothness ($\downarrow$)\\ (RRT)\end{tabular} & \begin{tabular}[c]{@{}c@{}}Smoothness ($\downarrow$)\\ (RRT+Refinement)\end{tabular} \\ \hline
0.00                                                      & $0.6017 \pm 0.0841$                                                                      & $0.5748 \pm 0.0778$                                                                                & $0.0236 \pm 0.0114$                                                                 & $0.0300 \pm 0.0135$                                                                            & $0.0124 \pm 0.0051$                                                                    & $0.0003 \pm 0.0003$                                                                               \\
0.25                                                                   & $0.5818 \pm 0.0723$                                                                      & $0.5477 \pm 0.0599$                                                                                & $0.0252 \pm 0.0121$                                                                 & $0.0345 \pm 0.0118$                                                                            & $0.0133 \pm 0.0049$                                                                    & $0.0003 \pm 0.0002$                                                                               \\
0.50                                                                   & $0.5685 \pm 0.0580$                                                                      & $0.5441 \pm 0.0465$                                                                                & $0.0225 \pm 0.0086$                                                                 & $0.0311 \pm 0.0115$                                                                            & $0.0136 \pm 0.0058$                                                                    & $0.0004 \pm 0.0003$                                                                               \\
0.75                                                                   & $0.6390 \pm 0.1773$                                                                      & $0.5875 \pm 0.1298$                                                                                & $0.0267 \pm 0.0112$                                                                 & $0.0345 \pm 0.0133$                                                                            & $0.0134 \pm 0.0047$                                                                    & $0.0003 \pm 0.0003$                                                                               \\
1.00                                                                   & $0.6048 \pm 0.0806$                                                                      & $0.5590 \pm 0.0636$                                                                                & $0.0301 \pm 0.0122$                                                                 & $0.0360 \pm 0.0141$                                                                            & $0.0151 \pm 0.0058$                                                                    & $0.0003 \pm 0.0003$                                                                               \\
1.25                                                                   & $0.6643 \pm 0.1626$                                                                      & $0.6084 \pm 0.1160$                                                                                & $0.0286 \pm 0.0112$                                                                 & $0.0334 \pm 0.0165$                                                                            & $0.0158 \pm 0.0074$                                                                    & $0.0003 \pm 0.0003$                                                                               \\
1.50                                                                   & $0.6772 \pm 0.1912$                                                                      & $0.6275 \pm 0.1575$                                                                                & $0.0330 \pm 0.0100$                                                                 & $0.0378 \pm 0.0142$                                                                            & $0.0136 \pm 0.0074$                                                                    & $0.0003 \pm 0.0003$                                                                               \\
1.75                                                                   & $0.7822 \pm 0.2598$                                                                      & $0.7031 \pm 0.2138$                                                                                & $0.0358 \pm 0.0072$                                                                 & $0.0440 \pm 0.0073$                                                                            & $0.0173 \pm 0.0066$                                                                    & $0.0002 \pm 0.0002$                                                                               \\
2.00                                                                   & $0.7721 \pm 0.2593$                                                                      & $0.7072 \pm 0.2272$                                                                                & $0.0392 \pm 0.0093$                                                                 & $0.0459 \pm 0.0091$                                                                            & $0.0147 \pm 0.0057$                                                                    & $0.0002 \pm 0.0002$                                                                               \\ \hline
\end{tabular}
}
\end{table*}

\begin{table}[b]
\caption{Planning conservatism vs scenes with increasingly larger Gaussian primitives.}
\label{tab:exp1-tab2}
\resizebox{\linewidth}{!}{
    \begin{tabular}{lcc}
    \hline
    Scene & Path Length ($\downarrow$) & Closest Distance ($\uparrow$) \\ \hline
    $M_1$ (Smallest Gaussians) & $0.6048 \pm 0.0806$        & $0.030 \pm 0.012$             \\
    $M_2$ & $0.6747 \pm 0.1721$        & $0.031 \pm 0.014$             \\
    $M_3$ (Largest Gaussians) & $0.7531 \pm 0.2812$        & $0.035 \pm 0.013$             \\ \hline
    \end{tabular}
}
\end{table}

\subsubsection{Linearization Effect}
\label{sec:experiments-linearize}
\begin{figure}[t]
    \begin{center}
        \includegraphics[width=0.9\linewidth]{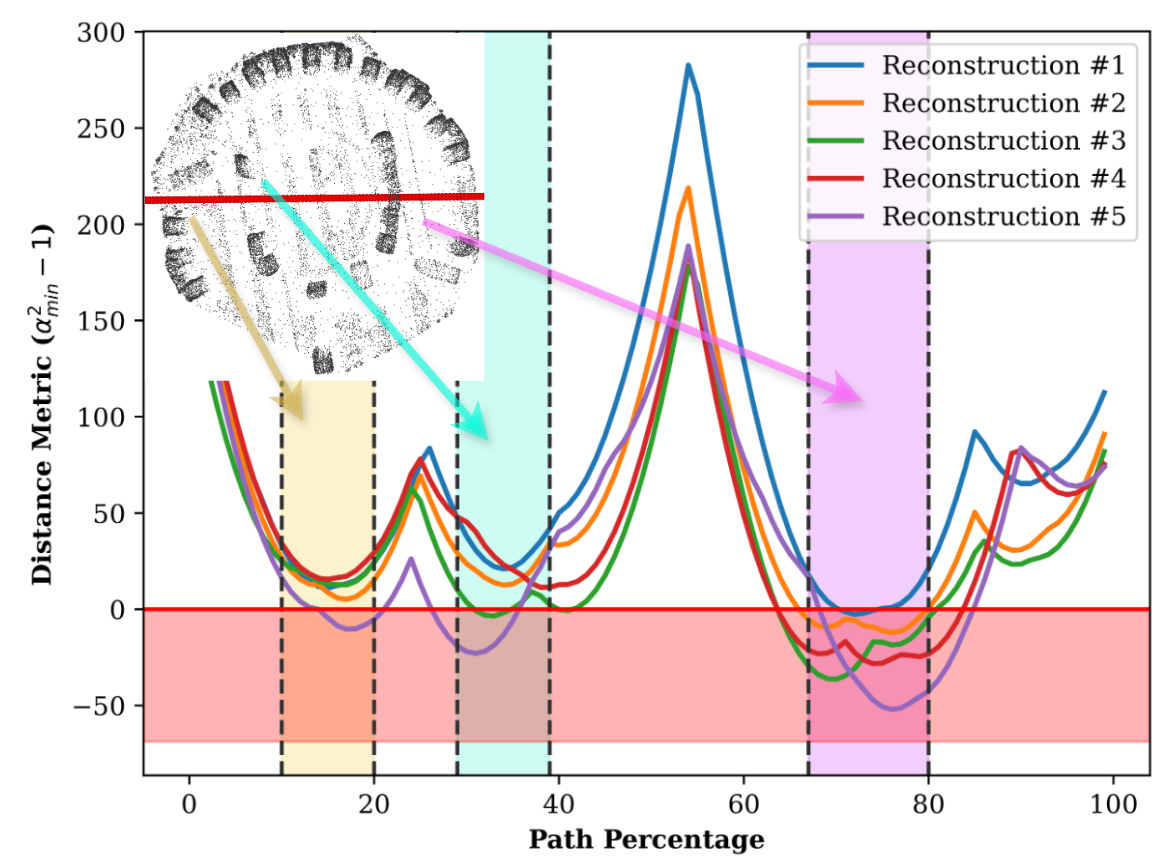}
        \caption{Computed collision distance along a linear path crossing reconstructed scenes with different Gaussian scales. As the distribution of large Gaussians increases, the metric yields smaller (more conservative) distance values, especially at sections of the path that traverse across obstacles.}
        \label{fig:distance_along_path}
    \end{center}
\end{figure}
As discussed in Sec.~\ref{sec:distance_formulation}, uncertainty propagation via linearization in Eq.~\ref{eq:linearization} is accurate when Gaussian scales are sufficiently smaller than the robot size (i.e., the minimum distance). In this section, we study our metric's response as the Gaussian-to-robot scale ratio increases. According to Eq. \ref{eq:linearization}, the propagated scaling uncertainty grows proportionally with the Gaussian scale $\Sigma_i$. Consequently, in poorly observed or uncertain regions characterized by large Gaussians, our metric inherently overestimates the collision risk. This is a highly desirable property, as it encourages more conservative behavior in regions of the map that are more uncertain. To validate this, we train five 3DGS Stonehenge models with increasing densification thresholds to generate models with increasing Gaussian sizes (M1--M5). First, we compute our distance metric along a straight collision-free path and plot the output for each model in Fig.~\ref{fig:distance_along_path}. As shown, for scenes with larger Gaussians, our distance metric produces smaller values, indicating more conservativeness. Next, we run the planner discussed earlier on three different scenes and report the result in Table~\ref{tab:exp1-tab2}. As expected, we observe that the planner operates more conservatively in scenes with larger Gaussians. 

\begin{figure*}[t]
    \centering
    \subfigure[Collision-aware visual target reaching.]{
        \includegraphics[width=0.75\textwidth]{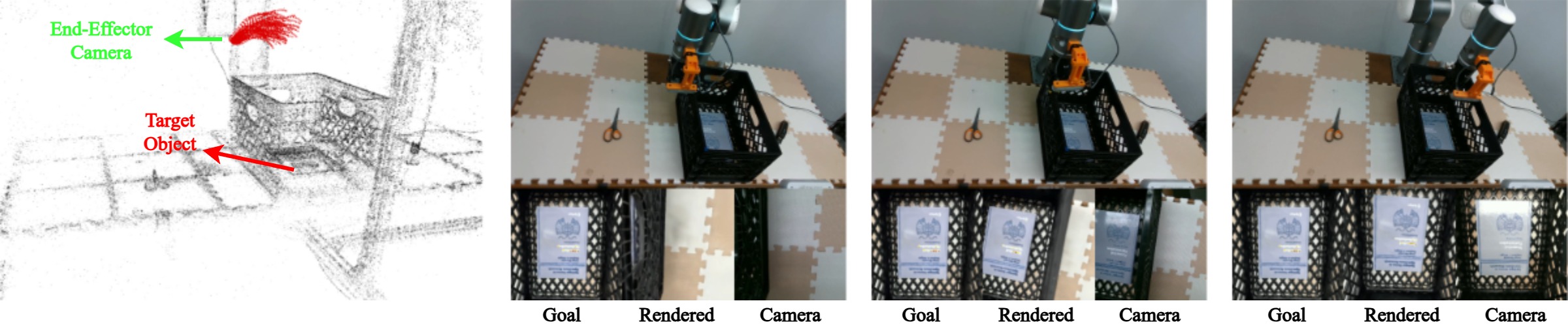}
        \label{fig:real-demos-reaching}
    }
    \subfigure[Collision-free path finding to image-defined objects.]{
        \includegraphics[width=0.75\textwidth]{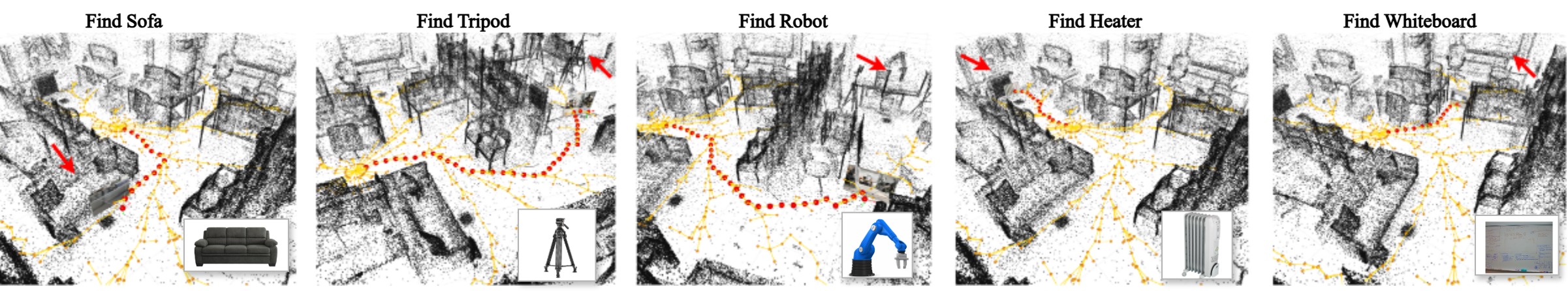}
        \label{fig:real-demos-semantic-rrt}
    }
    \subfigure[Collision-aware goal navigation.]{
        \includegraphics[width=0.75\textwidth]{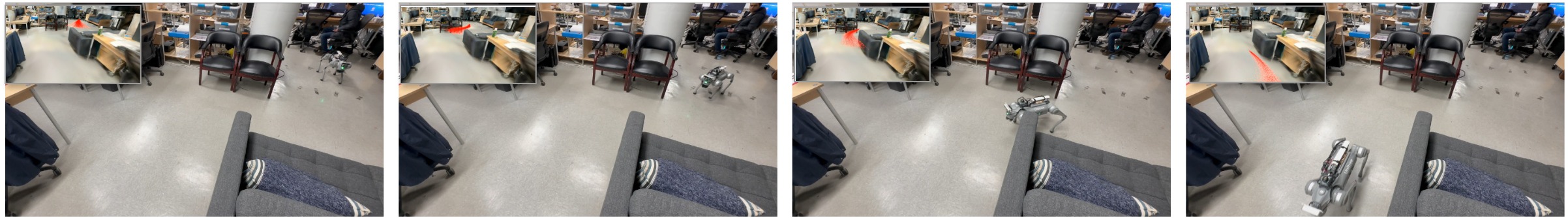}
        \label{fig:real-demos-navigation}
    }
    \caption{CollisionSplatting deployed to real-world locomotion and manipulation for collision-aware visual reaching and object navigation tasks.}
    \label{fig:real-demos}
\end{figure*}

\section{Real-World Demonstrations}
\label{sec:demonstrations}
\subsection{Joint Visual-Geometric Planning in Manipulation}
Our efficient distance metric, along with the real-time rendering capability of the 3DGS rasterizer, enables us to formulate control problems in which geometric terms, such as collision avoidance, are jointly optimized with image-based objectives. In this section, we demonstrate one example of such a formulation in a visual goal-reaching manipulation task. 
\subsubsection{Task and Setup}
We consider a Flexiv Rizon10s 7-DoF manipulator robot equipped with a wrist-mounted camera and tasked with reaching a desired target image while avoiding collisions with surrounding obstacles. The experimental setup is shown in Fig.~\ref{fig:real-demos-reaching}. 
\subsubsection{Formulation}
We implement the MPPI controller described in Sec.~\ref{sec:mppi} for our robotic arm. Each link of the arm is encapsulated by an ellipsoidal primitive, whose poses are computed over the planning horizon using a custom GPU-accelerated forward kinematics Warp kernel. At each control update, a batch of images is synthesized for the terminal state of each MPPI rollout via the 3DGS rasterizer. These images are fed to the visual reward model to compute the semantic cost (in the DINOv2 embedding space in our example). After a fixed number of updates, the first control output is sent to the robot, and the process is repeated with a warm start from the shifted solution from the previous step. Note that our highly efficient collision metric is essential for this example because while reactive collision avoidance may not require long-horizon MPC, proper long-horizon exploration is essential for optimizing the terminal visual reward without getting trapped in local minima.

\subsubsection{Result}
Table~\ref{table:exp-manipulation-quant} reports the reaching accuracy across five randomized goal images. Notably, optimizing purely for visual reaching accuracy without the collision cost results in paths that physically intersect the environment, leading to a $100\%$ collision rate with the box walls. 
% Therefore, while the unconstrained system nominally achieves lower translation errors, these trajectories are physically unexecutable. In contrast, enabling our CollisionSplatting cost successfully keeps the robot away from obstacles (maintaining a positive clearance of $0.017$ m), slightly sacrificing terminal visual precision to guarantee safe, collision-free execution. 
Thus the unconstrained system's lower translation errors are physically unexecutable. Enabling our CollisionSplatting cost keeps a positive clearance (0.017 m), trading slight visual precision for collision-free execution.
The planner is then deployed on the real robot, where the MPPI operates at 5 Hz over a receding horizon, and a subset of the resulting trajectory is executed by a compliant impedance controller at 1000~Hz. As qualitatively shown in Fig.~\ref{fig:real-demos-reaching}, the robot successfully reaches the goal image while avoiding contact with the environment.

\begin{table}[t]
\centering
\caption{Visual target reaching performance with or without collision cost.}
\label{table:exp-manipulation-quant}
\resizebox{\linewidth}{!}{
\begin{tabular}{lccc}
\toprule
\textbf{Configuration} & \textbf{Distance to} & \textbf{Translation Error} & \textbf{Rotation Error} \\ 
& \textbf{Closest Splat (m $\uparrow$)} & \textbf{(m $\downarrow$)} & \textbf{(degrees $\downarrow$)} \\ 
\midrule
Without Collision Cost & $-0.0212 \pm 0.029^{\dagger}$ & $\mathbf{0.0147 \pm 0.0057}$ & $\mathbf{3.4806 \pm 1.0627}$ \\
With Collision Cost & $\mathbf{0.017 \pm 0.008}$ & $0.031 \pm 0.014$ & $5.693 \pm 1.503$ \\ 
\bottomrule
\multicolumn{4}{l}{\footnotesize $^{\dagger}$ A negative distance indicates physical collision with the environment.}
\end{tabular}
}
\end{table}

\subsection{Image-Goal-Based Navigation}
In this example, we integrate the semantic DINOv2-based cost and our distance metric to formulate a dual-stage RRT+MPPI planner that enables collision-aware path planning toward visually specified targets. 
\subsubsection{Task and Setup}
In this task, a mobile robot (Unitree-Go2) with an internal 3DGS representation of the world is provided with a target image and must navigate to a semantically similar object within the environment.
\subsubsection{Formulation}
Instead of searching over the entire space randomly to find an object, we leverage our distance metric to execute a fast batched-RRT in exploration mode (without a predefined geometric target). This will rapidly generate a collection of collision-free robot configurations along with the paths to reach them. This greatly reduces the search space during image retrieval and ensures a feasible path from the current state to the retrieved image. During the retrieval phase, we synthesize the expected image observation for each node in the tree using the 3DGS rasterizer and compute the DINO feature embedding. Given the target image embedding, we select a semantically similar node in the tree. Finally, we use the MPPI formulation discussed earlier to control the robot along the path while avoiding obstacles on the fly. 
\subsubsection{Result}
As illustrated in Fig.~\ref{fig:real-demos-semantic-rrt}, the accelerated RRT plus the DINO distance enables collision-free path retrieval based on generic object goal-images, even when their visual appearances differ from the real scene. As shown in Fig.~\ref{fig:real-demos-navigation}, this path serves as a reference for our collision-aware MPPI controller. The controller then optimizes a shorter horizon of command velocities and sends the first command to the robot’s onboard locomotion policy at each MPC step. Notably, in this experiment, we provided only the final goal position found by the RRT to the MPPI controller, discarding intermediate waypoints. This serves to stress-test the controller’s collision-avoidance capability without dense geometric guidance from the global plan. Because our distance metric is highly efficient, we were able to extend the MPPI prediction horizon long enough to autonomously navigate around obstacles while maintaining real-time control frequencies.

\section{Conclusion}
\label{sec:conclusion}
% We presented CollisionSplatting, a simple GPU-friendly probabilistic distance metric for collision-aware planning directly in standard 3D Gaussian Splatting scenes. The metric propagates each Gaussian’s spatial uncertainty through an ellipsoid-based scaling-function distance, exposing a single tunable conservatism parameter that can be adjusted to meet desired safety margins. Leveraging its feed-forward, streaming structure, we integrated this metric as a batched collision-checking primitive in both a GPU-accelerated RRT and an MPPI controller, enabling joint optimization of geometric safety costs and image-conditioned objectives via rasterized 3DGS observations and DINO-style embeddings. In simulation, we demonstrated competitive or better collision-classification performance relative to Splat-Nav, SAFER-Splat, ATLASNav, and SPLANNING while substantially improving throughput and reducing VRAM requirements, and analyzed how conservatism scales with Gaussian size and reconstruction quality. Finally, we demonstrated collision-aware visual reaching and image-goal navigation on real robot platforms, showing how 3DGS combined with CollisionSplatting can serve as a practical interface between rich perceptual representations and real-time motion planning and control.

We presented CollisionSplatting, a GPU-friendly probabilistic distance metric for collision-aware planning directly in standard 3DGS scenes. The metric propagates each Gaussian's spatial uncertainty through an ellipsoid-based scaling-function distance, exposing a single tunable conservatism parameter. Owing to its feed-forward, streaming structure, we integrated it as a batched collision-checking primitive in both a GPU-accelerated RRT and an MPPI controller, jointly optimizing geometric safety and image-conditioned objectives via rasterized 3DGS observations and DINO embeddings. In simulation, our metric matched or exceeded Splat-Nav, SAFER-SPLAT, ATLASNav, and SPLANNING while substantially improving throughput and VRAM use. On real robots, we demonstrated collision-aware visual reaching and image-goal navigation, showing 3DGS with CollisionSplatting as a practical interface between rich perception and real-time control.
\bibliographystyle{IEEEtran}
\bibliography{references}
\end{document}